\documentclass[sigconf]{acmart}

\usepackage{subcaption}
\AtBeginDocument{%
  }

\copyrightyear{2025}
\acmYear{2025}
\setcopyright{cc}
\acmConference[ICPP Companion '25]{54th International Conference on Parallel Processing Companion}{September 08--11, 2025}{San Diego, CA, USA}
\acmBooktitle{54th International Conference on Parallel Processing Companion (ICPP Companion '25), September 08--11, 2025, San Diego, CA, USA}
\acmDOI{10.1145/3750720.3757281}
\acmISBN{979-8-4007-2109-0/25/09}

\newcommand{\hong}[1]{#1}

\begin{document}

\title{AI Assistants to Enhance and Exploit the PETSc Knowledge Base}

\author{Barry Smith}
\affiliation{%
  \institution{Flatiron Institute, Simons Foundation}
  \city{New York}
  \state{NY}
  \country{USA}
}
\email{bsmith@petsc.dev}

\author{Junchao Zhang}
\affiliation{%
  \institution{Argonne National Laboratory}
  \city{Lemont}
  \state{IL}
  \country{USA}
}
\email{jczhang@anl.gov}

\author{Hong Zhang}
\affiliation{%
  \institution{Argonne National Laboratory}
  \city{Lemont}
  \state{IL}
  \country{USA}
}
\email{hongzhang@anl.gov}

\author{Lois Curfman McInnes}
\affiliation{%
  \institution{Argonne National Laboratory}
  \city{Lemont}
  \state{IL}
  \country{USA}
}
\email{curfman@anl.gov}

\author{Murat Ke\c{c}eli}
\affiliation{%
  \institution{Argonne National Laboratory}
  \city{Lemont}
  \state{IL}
  \country{USA}
}
\email{keceli@anl.gov}

\author{Archit Vasan}
\affiliation{%
  \institution{Argonne National Laboratory}
  \city{Lemont}
  \state{IL}
  \country{USA}
}
\email{avasan@anl.gov}

\author{Satish Balay}
\affiliation{%
  \institution{Argonne National Laboratory}
  \city{Lemont}
  \state{IL}
  \country{USA}
}
\email{balay@anl.gov}

\author{Toby Isaac}
\affiliation{%
  \institution{Argonne National Laboratory}
  \city{Lemont}
  \state{IL}
  \country{USA}
}
\email{tisaac@anl.gov}

\author{Le Chen}
\affiliation{%
  \institution{Argonne National Laboratory}
  \city{Lemont}
  \state{IL}
  \country{USA}
}
\email{le.chen@anl.gov}

\author{Venkatram Vishwanath}
\affiliation{%
  \institution{Argonne National Laboratory}
  \city{Lemont}
  \state{IL}
  \country{USA}
}
\email{venkat@anl.gov}

\renewcommand{\shortauthors}{Smith et al.}

\begin{abstract}
Generative AI, especially through large language models (LLMs), is transforming how technical knowledge can be accessed, reused, and extended. 
PETSc, a widely used numerical library for high-performance scientific computing, has accumulated a rich but fragmented knowledge base over its three decades of development, spanning source code, documentation, mailing lists, GitLab issues, Discord conversations, technical papers, and more. Much of this knowledge remains informal and inaccessible to users and new developers. 
To activate and utilize this knowledge base more effectively, the PETSc team has begun building an LLM-powered system that combines PETSc content with custom LLM  tools—including retrieval-augmented generation (RAG), reranking algorithms, and chatbots—to assist users, support developers, and propose updates to formal documentation.
This paper presents initial experiences designing and evaluating these tools, focusing on system architecture, using RAG and reranking for PETSc-specific information, evaluation methodologies for various LLMs and embedding models, and user interface design. Leveraging the Argonne Leadership Computing Facility resources, we analyze how LLM responses can enhance the development and use of numerical software, with an initial focus on scalable Krylov solvers. Our goal is to establish an extensible framework for knowledge-centered AI in scientific software, enabling scalable support, enriched documentation, and enhanced workflows for research and development. We conclude by outlining directions for expanding this system into a robust, evolving platform that advances software ecosystems to accelerate scientific discovery.
\end{abstract}

\ccsdesc[500]{Mathematics of computing~Solvers}
\ccsdesc[500]{Information systems~Language models}
\ccsdesc[500]{Software and its engineering~Maintaining software}

\keywords{PETSc, AI, LLM, RAG, Knowledge base}

\maketitle

\section{Introduction}

\begin{figure}[htbp]
\centering
\hspace{-0.15in}\includegraphics[width=0.7\linewidth]{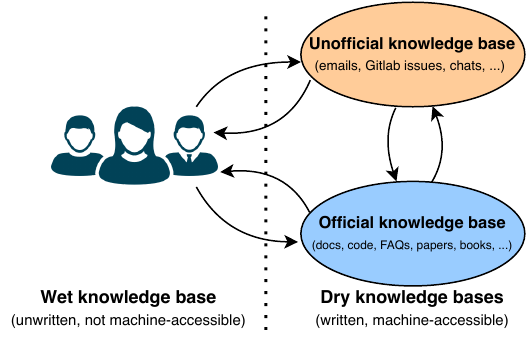}
\caption{PETSc community knowledge base. AI assistants could help the flow of knowledge in all directions.}
\label{fig:knowledge}
\end{figure}

High-performance scientific computing—which underpins discovery and innovation in all areas of science, engineering, technology, and society—relies on complex, evolving software libraries and tools that demand deep domain knowledge to develop, use, and maintain across continual advances in science drivers and computing architectures. 
Generative AI provides unprecedented opportunities to transform strategies for research and development (R\&D) of scientific software, with the ultimate goal of accelerating scientific insight~\cite{DOE-WorkshopReportAI4Science2023,sssdu-workshop-report2023, NAIRR-software-workshop}. 
PETSc—the Portable, Extensible Toolkit for Scientific Computation—has served the scientific computing community for over three decades as a foundational software infrastructure for scalable numerical solvers~\cite{petsc-user-ref,petsc-ecp2025,petsc-community2022}. 
During this time, the project has cultivated a sophisticated and robust code base and a deep, distributed body of developer and user knowledge.
The \textit{community knowledge base} includes source code and documentation, email exchanges in public mailing lists, GitLab issues, git merge request reviews, Discord chats, manual pages and examples from the PETSc website, written materials such as papers and presentations, and importantly,
PETSc developers and users' understanding of PETSc. Among the written material, the structured and curated subset constitutes the \textit{official} knowledge base:  documentation, code, frequently asked questions (FAQs), and other resources that have undergone refinement and review, essentially everything that has passed a git merge request, while the remaining is the \textit{unofficial} knowledge base.
We call the subset of community knowledge that is available only in a developer's brain (i.e., \textit{not written} in a machine-accessible format) the \textit{wet} knowledge base\footnote{\url{https://en.wikipedia.org/wiki/Wetware_(brain)}}, and the opposite the \textit{dry} knowledge base,
see Fig.~\ref{fig:knowledge}.

Historically, the PETSc knowledge base has grown through the daily work of developers and the continuous stream of user interactions. 
Because this content is implicit—captured in natural language, scattered across platforms, and often inaccessible without deep familiarity—this 
knowledge is only partially utilized. 
Developers may recall past issues vaguely, struggle to search 20-year-old mailing list archives, or repeatedly answer similar questions without updating the documentation. Similarly, users of PETSc may encounter steep learning curves despite abundant (but fragmented) resources, since any PETSc-related information and data can serve as a source of knowledge for developers and users.

Recent advances in generative AI, particularly large language models (LLMs), open a new chapter in leveraging knowledge bases. 
With LLMs’ ability to retrieve, synthesize, and contextualize natural language content, we can now utilize knowledge bases in transformative ways. 
In this framing, knowledge bases can be considered a reservoir, and LLM components—retrieval-augmented generation (RAG), reranking algorithms, Discord-integrated bots, and more—act as the pipes, filters, and faucets that deliver timely, accurate, and contextualized information to users and developers. 

These tools do not replace human contributors but act as scalable collaborators—suggesting code, summarizing discussions, generating documentation updates, and responding to user questions while adhering to PETSc’s collaborative development process. These same tools and techniques can and will be used to transfer material from the unofficial knowledge base to the official knowledge base, for example, dramatically improving manual pages with  little developer effort. 
Developer-user (as well as developer-developer) communication is a common way that information is transferred from the wet knowledge base to machine-accessible format. 
Questions motivate developers to write down (often informally) knowledge and conceptualizations that they  have never written down before. By having the LLM process these developer-user interactions, we can effectively capture wet knowledge as the developers perform their daily tasks without requiring them to try to core-dump their wet knowledge actively.

This paper presents initial efforts to develop LLM-based tools that enhance and exploit the PETSc knowledge base.
\begin{itemize}
    \item To enhance the PETSc knowledge base, we are improving the flow of knowledge (from wet to dry and from unofficial to official)—resulting in more information about effectively using PETSc being vetted and readily accessible, both directly to users and also to LLMs.
    \item To exploit the PETSc knowledge base, we are creating AI assistants that serve as knowledgeable, tireless partners in the daily workflows of PETSc R\&D and user support—that is, a system where AI tools are dynamic connectors between broad technical content and diverse needs of users and developers.
\end{itemize}

\begin{figure}[htbp]
\centering
\includegraphics[width=.99\linewidth]{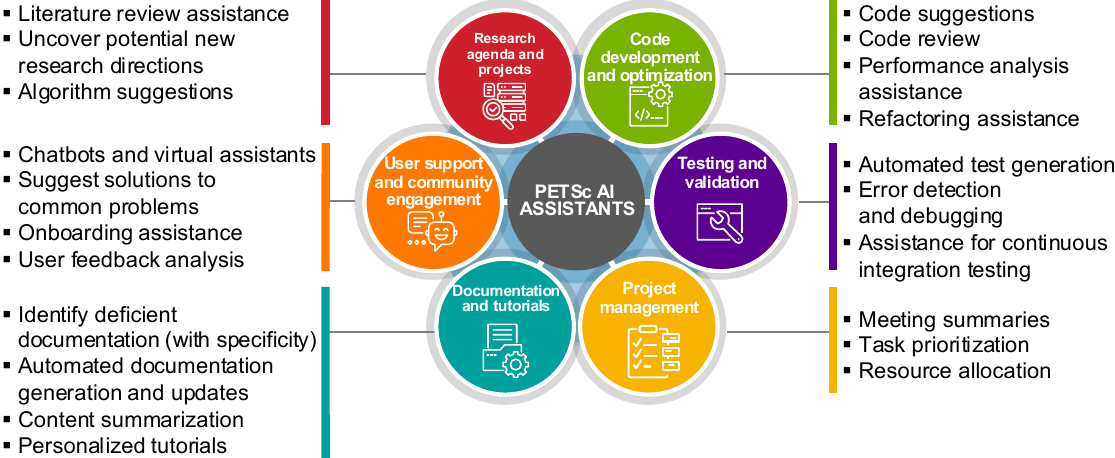}
\caption{Our goal is to create PETSc AI assistants capable of understanding and executing scientific computing tasks at the heart of numerical library research, providing the AI equivalent of human PETSc developers using LLM technology, while also continually advancing the PETSc knowledge base.}
\label{fig:petsc-ai-topics}
\end{figure}
Fig. \ref{fig:petsc-ai-topics} illustrates six key domains in which we envision LLM-powered assistants to meaningfully assist PETSc R\&D and user support while also augmenting the PETSc knowledge base for long-term community advances:
\paragraph{\bf User support and community engagement} Assistants embedded in tools such as Discord can answer user queries, recommend solutions, aid onboarding around PETSc's broad functionality, and even analyze recurring feedback to guide improvements. 
For PETSc, these capabilities help address questions on the development of applications that build on PETSc scalable linear, nonlinear, and timestepping solvers as well as lower-level vector and matrix data structures for heterogeneous computing architectures~\cite{petsc-sf2022} and outer-loop numerical optimization and analytics. 
\paragraph{\bf Documentation and tutorials} LLMs can detect outdated or unclear documentation, generate summaries, and draft updates. Personalized tutorials tailored to a user’s needs or application area could dramatically lower the barrier to entry for new users.
For the PETSc community, such custom assistance could meet users where they are (e.g., helping to translate between the terminology used in scientific application areas and that employed by numerical algorithms researchers). 
\paragraph {\bf Code development and optimization} AI assistants can help in code review, suggest code improvements, aid in API design, offer performance tuning strategies, assist with refactoring complex code, and develop completely new functionality. 
Such assistants could help with continual advances needed in the PETSc code base to achieve efficient performance and address next-generation science challenges on ever-changing computer architectures.   
\paragraph {\bf Testing and validation} LLMs can help generate automated tests, detect bugs, and assist in debugging or improving continuous integration practices. Such capabilities would help
open-source libraries like 
PETSc achieve comprehensive, multilevel testing to ensure a robust and reliable code base as part of the broader scientific software ecosystem.  
\paragraph{\bf Research agenda and projects} Beyond software development and maintenance, LLMs can support scientific discovery itself—by aiding in literature reviews~\cite{antu2023using}, surfacing unexplored research directions~\cite{schmidgall2025agent}, and suggesting algorithms for novel problems. 
\paragraph {\bf Project management} Through meeting summarization, task prioritization, and resource tracking, AI tools can reduce cognitive load and coordination overhead among contributors. Such capabilities help with communication and synchronization across the multi-institutional distributed PETSc developer team.

We do not aim to replace human developers, but rather to extend their capabilities and improve productivity. 
These AI assistants will operate within the collaborative infrastructure of PETSc—subject to the same review and git merge request processes as human contributors—to ensure that any changes to the official knowledge base are carefully vetted. 
Importantly, we recognize that LLMs are not infallible. Just as human developers occasionally submit flawed code or advice, so too will AI assistants. 
However, the PETSc community culture has robustly accommodated such imperfections through iterative improvement for many years. Thus, it is well suited to integrate these new tools while maintaining quality and trust.

Leveraging LLMs effectively within scientific computing requires more than generic chatbot capabilities. 
We need specialized techniques for corpus preparation, retrieval-augmented generation, fine-tuning, ranking, and continuous evaluation—tailored to scientific software development's technical depth, precision, and collaborative workflows.

Nonetheless, this endeavor is not without risk. The LLM landscape is dominated by proprietary systems with rapidly evolving capabilities. 
We aim to provide the best help for the PETSc user and developer community over time, enabling advanced scientific simulations that leverage emerging LLM capabilities. Consequently, our emphasis is on developing a framework for exploring PETSc AI assistants, which will provide practical capabilities for our community and enable us to focus on benchmarks and evaluation efforts that leverage our expert knowledge in numerical computing while contributing to improvements in the overall LLM-powered PETSc knowledge base.

We have outlined a vision of how LLMs can help the PETSc community; it is impossible to fully describe all aspects of our current and future work in this short manuscript. 
This paper presents two of our initial steps toward realizing this vision: (1) a RAG pipeline enhanced with reranking to improve retrieval from the PETSc knowledge base and (2) a PETSc chatbot embedded in Discord that enables real-time, AI-assisted conversations between users and developers on specific PETSc user situations. These components are supported by processing scripts, prompt libraries, and agentic memory systems that track and evaluate interactions over time. Together, they represent our first steps toward a sustainable, AI-driven platform for capturing and using the collective knowledge embedded in the PETSc ecosystem.  

\section{PETSc User Support}
\label{sec:petsc-user-support}

User support, which we have consistently provided for over thirty years, is a fundamental responsibility of the PETSc team—as it is for any open-source scientific software project—to disseminate knowledge and realize the full impact of the software. 
Consequently, user support is our first focus when exploiting the PETSc knowledge base with AI assistants.
Like many scientific computing libraries, PETSc has been funded primarily  by the U.S. government. However, not only has this funding been limited, but it also has largely focused on supporting research in numerical computing, leaving few resources available for software maintenance and user support.
Over the years, PETSc has developed a large and diverse user base, with tens of thousands of users across dozens of scientific domains. As a result, user support has become a significant undertaking. For instance, in May 2025 alone, there were over 300 messages across PETSc mailing lists, Discord channels, and GitLab issues. While PETSc developers have remained deeply committed to supporting users, the volume and complexity of user inquiries can often overwhelm a small team, particularly when resolving issues that involve extended, multiday, or multiweek discussions.
Thus, to ensure that this process remains sustainable in the long term, we are particularly interested in leveraging recent advances in AI to assist with user support.

Although PETSc users can employ general-purpose AI tools such as ChatGPT, these tools can produce incorrect or hallucinated responses when answering technical questions about PETSc. The reason is primarily that mainstream LLMs do not have access to sufficient PETSc information. 
For example, in January 2025, we posed a PETSc-specific question to ChatGPT-4o and observed such limitations firsthand.
\normalsize{
\\
\textit{\textbf{Question}: What does KSPBurb do?}
\\
\textit{\textbf{ChatGPT}: KSPBurb is an implementation of a Krylov subspace method in
PETSc used to solve systems of linear equations. Specifically, KSPBurb is a block version of the unpreconditioned Richardson iterative method ...
}
}

We scored this response as 0:  it was a hallucination to a question about ``KSPBurb" -- a fictitious term that follows the 
naming convention of PETSc Krylov solvers (see Table \ref{tab:rubric} for the rubric used in the assessment).
Although later versions of ChatGPT gave better results, this example shows the risk of relying solely on mainstream LLMs.
The example also reminds us of the dangers of incorporating AI into user support. As noted in \cite{cursor}, a prominent platform experienced backlash when its AI agents, masquerading as human support staff, fabricated a nonexistent login policy in response to user inquiries that led to widespread account cancellations. To safeguard the integrity and reputation of PETSc, we initially have adopted a cautious approach, and to date we have avoided full automation in user support interactions.

A typical interaction between PETSc users and developers is highly interactive and rarely concludes with a single question-and-answer exchange. A user's initial question may be vague, lack context, or use incorrect PETSc terminology. In response, a developer will often ask for clarification. Other developers join the conversation, contributing new perspectives or identifying misunderstandings. As the discussion progresses, the user may pose follow-up questions that are now more precise and well formulated. Code snippets or test cases are sometimes shared, which should ideally be tried by the participants. This process can be likened to a treasure hunt, where context and clues gradually emerge and occasional experimentation is required to determine what works. Both users and developers may inadvertently provide incorrect information, so our process must be robust to identify and overcome these mistakes.

To address the above challenges, we propose in Section \ref{sec:petsc-ai-approach} an augmented PETSc LLM workflow that can effectively process and deliver PETSc-centric information, and in 
Section \ref{sec:discord-bots} we present an integrated LLMs and PETSc Discord bots system that can reduce the risk of hallucinated responses reaching users. 

\section{The PETSc Augmented LLM Workflow}

\label{sec:petsc-ai-approach}

\begin{figure}[htbp]
\centering
\includegraphics[width=1.0\linewidth]{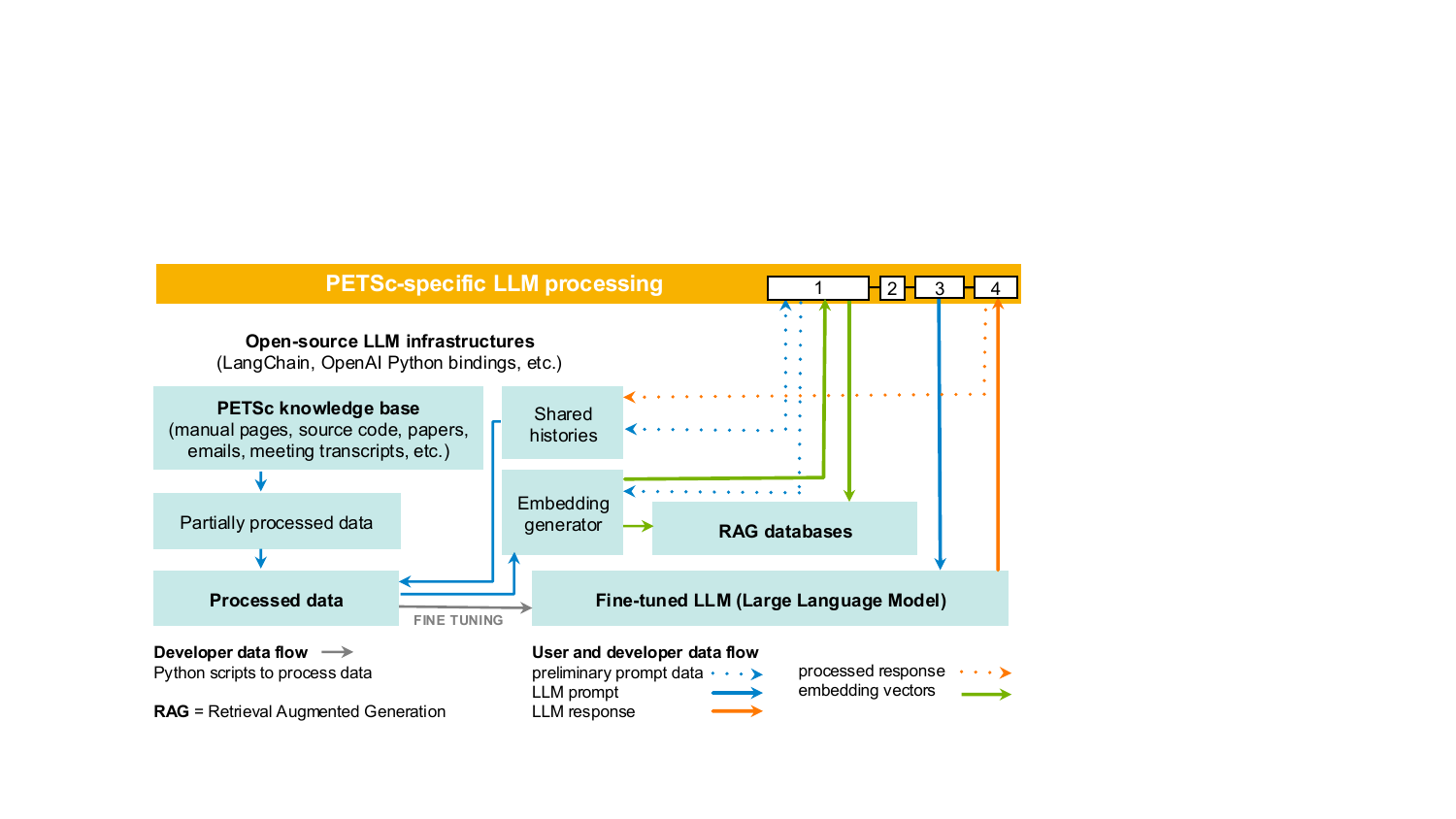}
\caption{Our approach for creating PETSc AI assistants builds on standard LLM ecosystem tools. The boxes  1, 2, 3, and 4 in the top bar indicate the sequence of processing for PETSc-specific LLM interactions.}
\label{fig:petsc-ai-approach}
\end{figure}

We have developed a preliminary PETSc LLM workflow, illustrated in Fig.~\ref{fig:petsc-ai-approach}, which 
can be used in multiple settings. For example, we could use the LLM to answer questions from the PETSc Discord server, from mailing lists, or from a chat box on the PETSc homepage. For developers, we could even provide command line tools and integrated development environment (IDE) extensions to facilitate various use cases.
Our approach builds on standard LLM tools, such as LangChain \cite{langchain}, while incorporating practical adaptations for PETSc. Specifically, we do the following:
\begin{itemize}
\item We use embedding methods, RAG, and reranking to tailor general-purpose LLM capabilities to PETSc-specific content.

\item We maintain a structured and detailed database of interactions, facilitating systematic improvements and comparisons between different LLMs and embedding approaches.

\item We provide developers with real-time access and control over the information presented to the LLMs, helping to ensure utility and accuracy.
\end{itemize}
The workflow is compatible with various continuation and embedding models, including state-of-the-art variants. In the following subsections we  describe PETSc-specific processing components in detail.

\subsection{Generating the RAG databases}

Retrieval-augmented generation algorithms ensure that users receive the most contextually appropriate and accurate information. RAG consists of two phases: the generation phase and the retrieval phase. Here, we discuss the generation phase.
PETSc documentation is maintained in Markdown format and processed via Sphinx \cite{sphinxdoc} into the website format. To construct a practical RAG database using this documentation, we use the following steps (see Fig.~\ref{fig:petsc-ai-approach}, beginning with ``PETSc knowledge base", following the blue arrows to ``Embedding generator" and then the green arrow to ``RAG databases"): 
\begin{itemize}
\item Employ LangChain's DirectoryLoader, UnstructuredMarkdownLoader, and  RecursiveCharacterTextSplitter.  
\item Input these results into Chroma.from\_documents to generate the vector database for a particular embedding (multiple databases can be built for different embeddings). 
\end{itemize}
These steps allow us to remove irrelevant content and embed direct reference links to relevant documentation, facilitating quick user references. Similar processes will be used for PETSc publications and the open PETSc mailing lists. Developers and users will be able to choose which
vector databases to use.

\subsection{Utilizing the documentation RAG database}

During retrieval, one encodes the user query into a vector numerical representation and searches in the RAG vector database for similar documents.
Then, the original user query is augmented by %
context information from the retrieved documents and sent to the LLM.
In Fig.~\ref{fig:petsc-ai-approach}, from box 1 follow the blue dotted arrow to the embedding generator, which returns the embedding vectors to box 1, where the RAG databases are searched and material is prepared to be passed to box 2.

\subsection{Augmenting RAG searches with PETSc-specific keyword searches}

We have augmented the RAG searches with PETSc-specific keyword searches. Whenever a word in the query has a PETSc manual page associated with it, for example, KSPSolve, the manual page is added to the material that RAG has found. In Fig.~\ref{fig:petsc-ai-approach}, not explicitly displayed, box 1 locates material from ``Processed data."

\subsection{Reranking enhanced RAG}

\begin{figure}
\center
\includegraphics[width=0.95\linewidth]{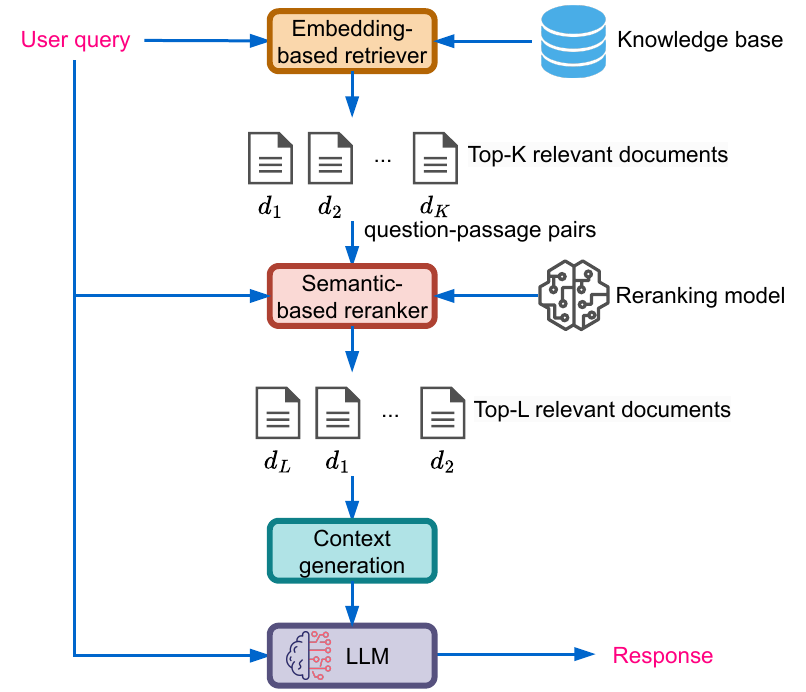}
\caption{Reranking-enhanced RAG workflow.}
\label{rerank}
\end{figure}

The embedding-based retrieval and keyword search prioritizes speed over accuracy since the knowledge databases
can be vast and complex.
Through vector search, the retriever quickly returns a few top results, which may include both relevant and tangential information.
\textit{Reranking} enhances retrieval by filtering and reordering retrieved documents according to refined relevance scores, reducing noise and irrelevant information. Reranking ensures that the LLM receives the most contextually appropriate information. The reranker refines the initial list of candidates,
using scoring models based on, for example, cross-encoders and specialized LLMs, moving the most contextually relevant documents to the top and possibly removing less relevant material completely. In Fig.~\ref{fig:petsc-ai-approach} this process occurs inside box 2 after the initial documents have been located in box 1;
Fig.~\ref{rerank} depicts this process in detail. In particular, in our studies discussed below, we generate $K=8$ candidate documents in the first pass and refine them down to $L=4$ documents with the reranker in the second pass. 

\subsection{Postprocessing LLM output for users}

The LLM output, depicted in Fig.~\ref{fig:petsc-ai-approach} with the orange arrow, is generally provided in Markdown. We (in box 4) provide tools that postprocess the Markdown before displaying it to users, such as converting it to HTML for display on a webpage. Itemized lists can be automatically detected and parsed.
In addition, we automatically detect blocks of code and can pass them to a compiler to verify that they work. 
We note that LLMs are now making it possible to return their output in JSON, making postprocessing easier since we do not have to ``reverse engineer" the LLM output. 

\subsection{Storing and tracking interaction history}

In addition to visually presenting the LLM output, we keep a detailed, manipulatable, searchable database of all  interactions with all the LLMs by the PETSc developers (and eventually also users). In Fig.~\ref{fig:petsc-ai-approach} this is depicted with the dotted orange arrow from box 4 to ``Shared histories"; the blue dotted line to box 1 indicates that material from the shared history will also eventually be included in the RAG and reranking processing and passed to the LLM. Questions, responses, dates when done, and the continuation and embedding model used for each interaction will be included, as well as the generated prompts in the interaction. We provide an easy way for ``scorers" to ``blind-score" the responses to questions and indicate correct and incorrect portions of the responses. We can also score answers from PETSc developers stored in the same database, establishing high quality and trust. Currently, we use a bespoke Python dictionary to manage the database, but we are pursuing using emerging agentic memory systems \cite{hatalis2023memory}.

\section{PETSc AI Bots in Discord}
\label{sec:discord-bots}

\begin{figure}
\centering
\includegraphics[width=0.9\linewidth]{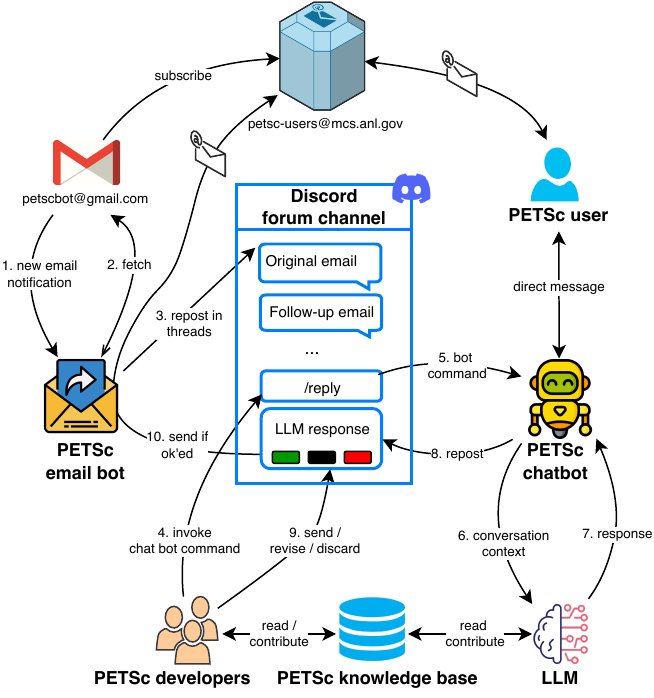}
\caption{Bots in Discord that connect a PETSc mailing list and an LLM. Numbered arcs show a typical sequence of events in the workflow.}
\label{fig:petsc-bots}
\end{figure}

Discord \cite{discord2025} is an instant messaging platform where communication happens in virtual communities called “servers.” 
We have set up a PETSc Discord server, and hundreds of users have joined since its
debut. We integrated augmented PETSc LLMs with the Discord server and set up a chatbot that has two usage modes: (1) one can directly chat with it, or (2) the chatbot can answer questions from emails, while PETSc developers can monitor and vet answers on back channels. 
In this section we first briefly introduce Discord's integration mechanisms, then describe the workflow that
connects a PETSc public mailing list and LLMs; see Fig.~\ref{fig:petsc-bots}.

Discord provides integrations via webhooks and applications (called apps or bots). 
Webhooks are user-defined HTTP callbacks. One can create webhooks and associate them with specific Discord channels. 
With the URL provided by Discord for a webhook, one can, for example, make an HTTP request to post a message to the associated channel. 
Discord users can set up Discord apps and associated bots. In turn, they are responsible for programming 
in Discord APIs (e.g., through Python) and eventually deploying their apps locally or in the cloud.
Apps could access messages, channels, and user accounts on Discord. 
They can provide commands for users to invoke or act as bots on Discord.

PETSc has three public mailing lists: \textit{\{petsc-users, petsc-maint, petsc-dev\}}. \textit{petsc-dev} is mainly for PETSc developers and others interested in PETSc development. 
The other two are for PETSc users with one distinction: \textit{petsc-maint} is a private maintenance email with controlled recipients and no public archives, 
while \textit{petsc-users} is fully public and has archives from the past twenty years. 
In this study we targeted \textit{petsc-users} but didn't touch its archives for RAG. 
We created a Google Gmail account \textit{petscbot@gmail.com} and subscribed it to the \textit{petsc-users} mailing list. 
Then, with Google Apps Script services \cite{googlescript}, we use JavaScript to periodically check whether there are new (unread) emails from \textit{petsc-users} in the Gmail account. 
If there are, the script sends a message to a webhook associated with a private channel named \textit{petsc-users-notification} in PETSc Discord. 
An email bot (application) was added to the PETSc Discord server, which watches for new messages in \textit{petsc-users-notification}. 
Whenever there is a new message, using Gmail's Python API, the email bot fetches unread emails from \textit{petsc-users} in the Gmail account and marks them read. 
The fetched emails are posted to a private channel, visible only to PETSc developers, named \textit{petsc-users-emails}. 
This channel is a Discord Forum channel made of posts. We can make each email thread a post, with its title being the email thread's subject 
and emails within the thread follow up messages in the post. We also fetch and attach email attachments to their corresponding messages in Discord. 
We lightly parse email bodies to remove quotes commonly seen in email replies and revert the url-defense protected URLs so that messages are presented concisely. 

From a post in \textit{petsc-users-emails}, any PETSc developer can invoke a command \texttt{/reply} on the chatbot, which will then build a conversation context with the title, messages, and their attachments of the post and present it to the LLM. 
The Discord to LLM connection code was based on llmcord \cite{llmcord} but was heavily adapted to our needs. 
The response from the LLM is added as a message to the post. All PETSc developers can see the message so that anyone can chime in. The message has three buttons: send, discard, and revise. If the response is fine, the developer clicks the send button to send the response to \textit{petsc-users} with a signature of the name of the developer who clicked the button.\footnote{The chatbot is still in private testing, and we haven't sent any LLM-generated responses to the petsc-users mailing list.}
Also, the message in Discord will be time-stamped and tagged with the developer's name. If the response is nonsense, developers can click the discard button to delete the message. Otherwise, if the response is partially correct and a developer clicks the revise button, the chatbot will ask the developer to guide a follow-up message. Developers can chat with the bot until a satisfying response is generated and ready to be sent. 
The Javascript for the Gmail account ignores emails from the chatbot and marks them as read, 
so that it doesn't repost messages in the channel.
In addition, users can do ``direct messages" with the chatbot, keeping their conversation private; however, this may expose the user to unvetted hallucinations of the LLM.

The above workflow leverages LLMs to help but not replace
PETSc developers answering user questions from the mailing list.
The negative example \cite{cursor} alerts us not to let 
misleading or hallucinated answers damage PETSc’s reputation in the community. 

\section{Preliminary Evaluation and Insights}
\label{sec:eval}

\subsection{Evaluation strategy}
To evaluate the quality of answers \hong{at the highest standard}, we leverage human expertise to classify the LLM responses using a blind-review workflow.
\hong{Two experienced PETSc developers (with 10+ years of experience) served as the reviewers; they}
assigned a single score for each
answer, based on correctness and completeness, without knowledge of
the process used to generate it.
The scoring rubric is presented in Table \ref{tab:rubric}.
We compiled a benchmark dataset comprising 37 questions on the use of Krylov methods within PETSc, widely used by diverse scientific applications as a key part of composable algebraic solvers~\cite{petsc-composable-solvers2012} and representative of a subtopic within the broader PETSc library. 
\hong{The questions were derived from real user inquiries and span a range of difficulty levels, including both easy and hard questions.
The hard questions are often open-ended and cannot be answered directly using the PETSc documentation.}

\begin{table}[h]
\begin{center}
\begin{tabular}{ | r l|} 
 \hline
 \textbf{0:} & Nonsensical answer\\ 
 \textbf{1:} & Incorrect or inaccurate statements (hallucinations) \\ 
 \textbf{2:} & Correct material with only minor inaccuracies \\
 \textbf{3:} & Answer is clear and correct \\
 \textbf{4:} & Ideal answer, close to what an expert would respond \\
 \hline
\end{tabular}
\caption{Rubric for LLM responses (higher is better).}
\label{tab:rubric}
\end{center}
\end{table}

\subsection{Comparison with LLM baseline}
We conducted experiments with several popular LLMs, including OpenAI's GPT-4 variants and Meta's Llama3 variants, alongside various embedding models.
Our analysis identified OpenAI's GPT-4o and
text-embedding-3-large as providing the best overall performance.
Consequently, we report results for this specific LLM
and embedding model combination.

We have explored the NVIDIA reranker (commercial) and the Flashrank reranker (free) through the LangChain interface.
The NVIDIA reranker provides GPU-accelerated transformer-based encoder models.
Flashrank provides lightweight models running on the CPU, including encoder (BERT-based) rerankers and non-encoder rerank models.
Both rerankers yield a similar level of accuracy for our database.
We  selected Flashrank in this study because of its speed.

Providing PETSc-specific information (via pretraining and RAG) enables us to overcome limitations in mainstream LLMs, which typically lack sufficient training data on PETSc to provide accurate and reliable technical responses. 
For instance, the example below demonstrates how our RAG system prevents the LLM hallucination regarding ``KSPBurb" presented above.
\normalsize{
\\
\textit{\textbf{Question}: What does KSPBurb do?}
\\
\textit{\textbf{ChatGPT}: It appears there may be a typo or misunderstanding, as there's no PETSc
function or object named KSPBurb. In PETSc, the KSP (Krylov Subspace) module
is used to solve linear systems iteratively and includes a variety of
solver types such as GMRES, CG, BiCG, among others ... 
}
}

\begin{figure}
\centering
\begin{subfigure}{\linewidth}
\includegraphics[width=0.99\linewidth]{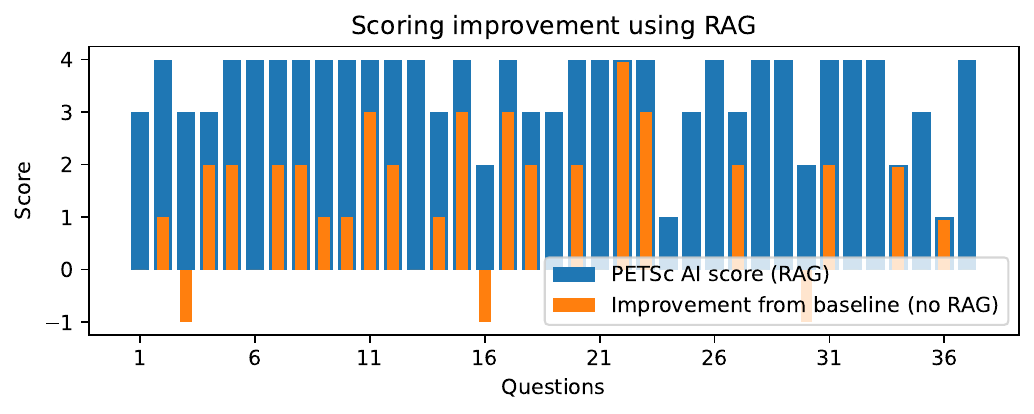}
\caption{}
\label{fig:score_rag}
\end{subfigure}
\begin{subfigure}{\linewidth}
\includegraphics[width=0.99\linewidth]{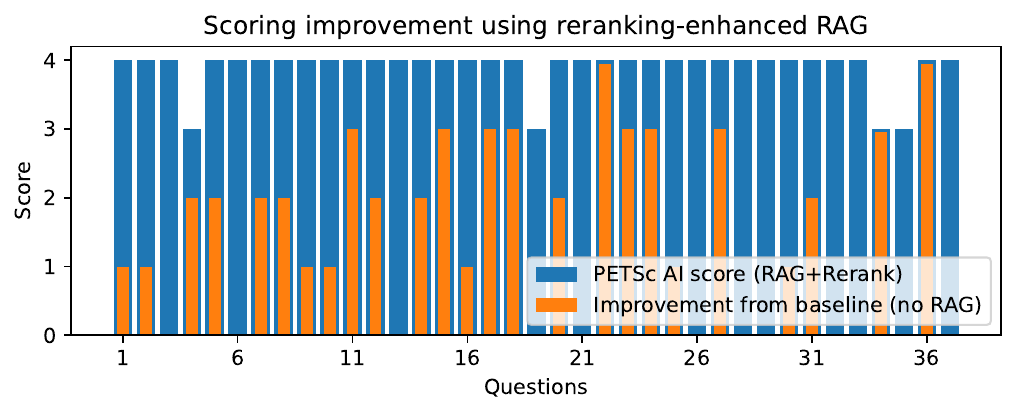}
\caption{}
\label{fig:score_ragrerank}
\end{subfigure}
\begin{subfigure}{\linewidth}
\includegraphics[width=0.99\linewidth]{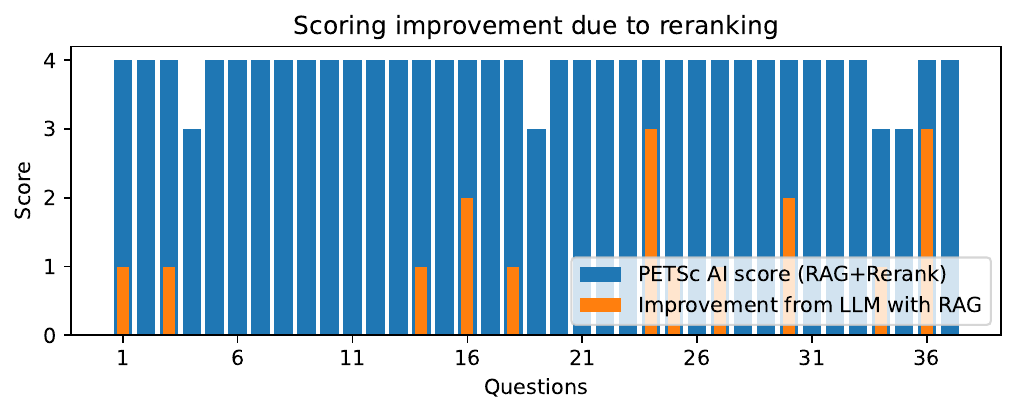}
\caption{}
\label{fig:score_rerank}
\end{subfigure}
\caption{Scoring comparison for LLMs.}
\end{figure}

Figs. \ref{fig:score_rag} and \ref{fig:score_ragrerank} 
present the evaluation scores for the benchmark questions, demonstrating an improvement over the baseline achieved by using RAG (Fig.~\ref{fig:score_rag}) or by using rerank-enhanced RAG (Fig.~\ref{fig:score_ragrerank}).
The baseline in both figures corresponds to using GPT-4o without RAG.  
As indicated by the orange bars in Fig.~\ref{fig:score_rag}, employing the LLM with RAG improves scores for 20 questions while negatively impacting scores for three questions.
Fig.~\ref{fig:score_ragrerank} demonstrates that utilizing the LLM with reranking-enhanced RAG improved scores for 25 questions, 
with no negative impact observed on any question's score.
This configuration achieved a perfect score (4) for 33  of 37 questions and a score of 3 for the remaining four questions.

\subsection{Impact of reranking}

Fig. \ref{fig:score_rerank}, which can also be inferred from Figs. \ref{fig:score_rag} and  \ref{fig:score_ragrerank},
illustrates the significant impact of the reranking algorithm on the RAG process.
The scores for 11 questions were improved by integrating reranking into the RAG process.
Two of these questions saw a score increase of 3 points!

\begin{figure}
\centering
\includegraphics[width=0.98\linewidth]{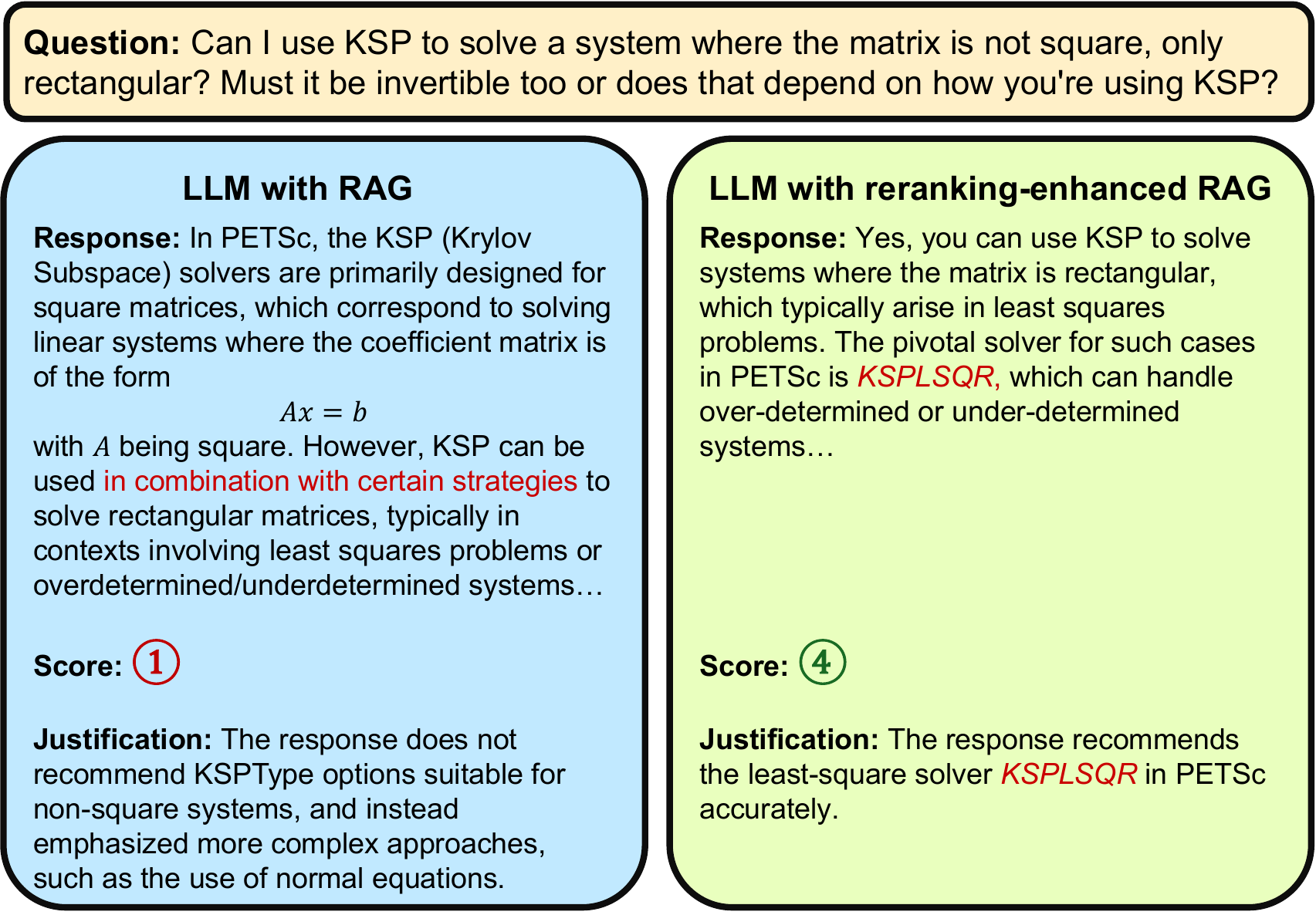}
\caption{Case Study 1: Comparison of RAG and reranking-enhanced RAG.}
\label{fig:case1}
\end{figure}
\begin{figure}
\centering
\includegraphics[width=0.98\linewidth]{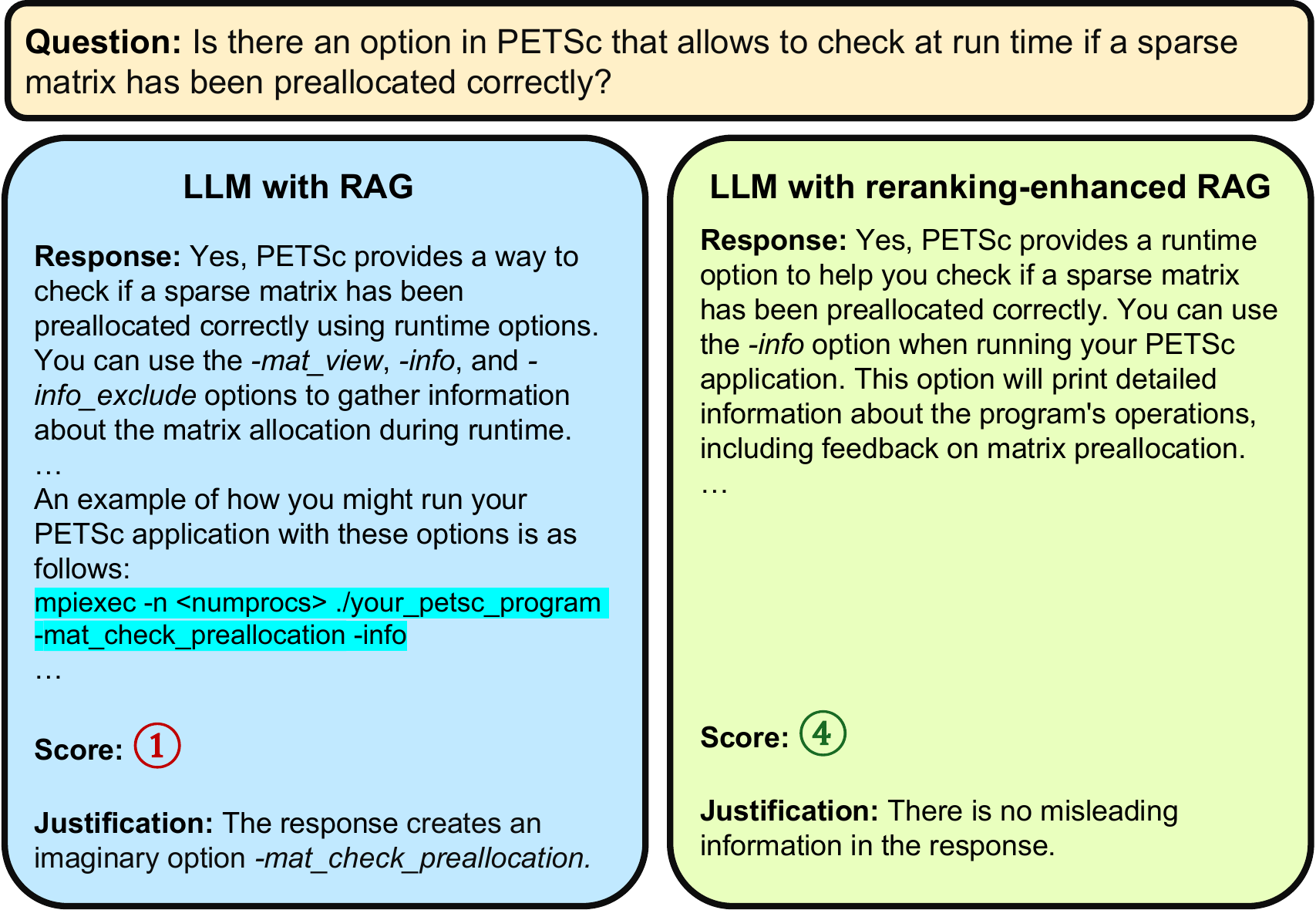}
\caption{Case Study 2: Comparison of RAG and reranking-enhanced RAG.}
\label{fig:case2}
\end{figure}

Figs. \ref{fig:case1} and  \ref{fig:case2} present the LLM responses and their scoring justifications for two illustrative case studies.
In the first case study, the RAG-enabled LLM failed to suggest the specific PETSc KSP solver for non-square systems.
However, the reranking-enhanced RAG successfully resolved this issue.
A closer investigation revealed that the reranking algorithm identified the following critical context, which was missed by the basic RAG algorithm:
\\
\textit{``KSP can also be used to solve least squares problems, using, for example, KSPLSQR ...''}

In the second case study, the LLM hallucinated an imaginary PETSc runtime option.
A comparison of contexts retrieved by the two RAG methods revealed only one common context, with the other three contexts being distinct.
The reranking-enhanced RAG, however, successfully retrieved a paragraph
discussing the \texttt{-info} option:
\\
\textit{``
As described above, the option -info will print
information about the success of preallocation
during matrix assembly ...''}\\
whereas the plain RAG algorithm did not find this information.

\subsection{Inference latency}
To evaluate inference latency, we measured separately the running time for the
RAG process and the LLM response.
This approach enabled us to quantify the overhead introduced by the RAG systems and assess their overall efficiency.
As presented in Table \ref{tab:timing}, adding the reranking stage increased the average RAG processing time by approximately $2.4$ times.
However, the average running time for the reranking-enhanced RAG remained less than 11\% of the average LLM response time.
\begin{table}
\renewcommand{\arraystretch}{1.2} %
    \begin{center}
        \small
        \begin{tabular}{ c c c c c c c}
        \toprule
            & \multicolumn{3}{c}{RAG} & \multicolumn{3}{c}{RAG+reranking} \\
            \cline{2-4}
            \cline{5-7}
            & Min & Max & Avg & Min & Max & Avg \\ 
            \midrule
            RAG time & 0.16 & 3.11 & \textbf{0.44} & 0.48 & 5.71 & \textbf{1.05} \\
            LLM response & 2.74 & 16.47 & 9.56 & 2.28 & 15.62 & 9.63 \\
\bottomrule
        \end{tabular}
        \caption{Run time for RAG and the LLM (in seconds) on an Intel i7-11700KF 3.6GHz CPU.}
        \label{tab:timing}
    \end{center}
\end{table}

\section{Related Work}
Recent advances in generative artificial intelligence, particularly in large language models, have significantly transformed software engineering. These models have demonstrated impressive capabilities in tasks such as code generation~\cite{chen2024ompgpt,nichols2024hpc, valerolara2023comparingllama2gpt3llms}, HPC software development~\cite{godoy2024large}, and compiler validation~\cite{munley2024llm4vv}. 
Building on these capabilities, RAG systems~\cite{lewis2020retrieval} have extended the use of LLMs by integrating external knowledge sources during inference. Several studies \cite{ali2024establishing, chen2023lm4hpc} have explored using RAG systems to support the development, maintenance, and transformation of legacy code bases. 

As a mature scientific computing library, PETSc contains much valuable but unstructured information, such as developer discussions and user Q\&As, that general-purpose LLMs are unlikely to have seen during training. This situation makes retrieval and formatting essential for effective use. Moreover, PETSc serves a diverse user base, so usability is critical. Furthermore, its focus on domain-specific, high-performance numerical computation requires knowledge that general LLMs typically lack, limiting their effectiveness without adaptation.

\section{Conclusion and Future Work}
\label{sec:conclusion}

Our preliminary results underscore the feasibility and significant potential of integrating LLMs (with PETSc-specific processing) to enhance the PETSc knowledge base and exploit it through AI assistants in daily workflows of PETSc research, development, and support. Future plans focus on advancing LLM-based tools specialized for PETSc and deepening their integration with PETSc activities to contribute to more robust and accessible scientific software environments, including exploring agentic approaches for complex tasks such as code generation and optimization.
We also want to incorporate additional information as part of PETSc-specific RAG and to put the PETSc Discord chatbot in production use.
The end result will be LLM assistants acting as and interacting daily with other PETSc developers, 
with goals of enhancing productivity and advancing scientific discovery. 

\begin{acks}
We thank the PETSc developers for discussing this work as well as for their many contributions to the PETSc code base and community knowledge base, without which this work would not be possible. 

This material is based upon work supported by Laboratory Directed Research and Development (LDRD) funding from Argonne National Laboratory, provided by the Director, Office of Science, of the U.S. Department of Energy (DOE) under Contract No. DE-AC02-06CH11357. 
Research was partially supported by the U.S. DOE Office of Science Distinguished Scientist Fellows Program.
This research used resources of the Argonne Leadership Computing Facility, a U.S. Department of Energy (DOE) Office of Science user facility at Argonne National Laboratory, and is based on research supported by the U.S. DOE Office of Science-Advanced Scientific Computing Research Program, under Contract No. DE-AC02-06CH11357.
\end{acks}

\bibliographystyle{ACM-Reference-Format}
\bibliography{refs,petsc-refs}

\end{document}